# Flexible Visual Quality Inspection in Discrete Manufacturing

Tomislav Petković, Darko Jurić and Sven Lončarić
University of Zagreb
Faculty of Electrical and Computer Engineering
Unska 3, HR-10000 Zagreb, Croatia
Email: {tomislav.petkovic.jr, darko.juric, sven.loncaric}@fer.hr

*Abstract*—Most visual quality inspections in discrete manufacturing are composed of length, surface, angle or intensity measurements. Those are implemented as end-user configurable inspection tools that should not require an image processing expert to set up. Currently available software solutions providing such capability use a flowchart based programming environment, but do not fully address an inspection flowchart robustness and can require a redefinition of the flowchart if a small variation is introduced.

In this paper we propose an acquire-register-analyze image processing pattern designed for discrete manufacturing that aims to increase the robustness of the inspection flowchart by consistently addressing variations in product position, orientation and size. A proposed pattern is transparent to the end-user and simplifies the flowchart. We describe a developed software solution that is a practical implementation of the proposed pattern. We give an example of its real-life use in industrial production of electric components.

## I. INTRODUCTION

Automated visual quality inspection is the process of comparing individual manufactured items against some pre-established standard. It is frequently used both for small and high-volume assembly lines due to low costs and non-destructiveness. Also, ever increasing demands to improve quality management and process control within an industrial environment are promoting machine vision and visual quality inspection with the goal of increasing the product quality and production yields.

When designing a machine vision system many different components (camera, lens, illumination and software) must be selected. A machine vision expert is required to select the system components based on requirements of the inspection task by specifying [1]:

1) **Camera**: type (line or area), field of view, resolution, frame rate, sensor type, sensor spectral range,
2) **Lens**: focal length, aperture, flange distance, sensor size, lens quality,
3) **Illumination**: direction, spectrum, polarization, light source, mechanical adjustment elements, and
4) **Software**: libraries to use, API ease of use, software structure, algorithm selection.

Selected components are then used to measure length, surface, angle or intensity of the product. Based on measurements a decision about quality of an inspected product is made. Although software is a small part of an quality inspection system usually, in addition to a machine vision expert, an image processing expert is required to select the software and to define the image processing algorithms that will be used.

Nowadays, introduction of GigE Vision [2] and GenICam [3] standards significantly simplified integration of machine vision cameras and image processing libraries making vision solutions for the industry more accessible by eliminating the need for proprietary software solutions for camera interaction. Existing open source image processing libraries especially suitable for development of vision applications, such as OpenCV (Open Source Computer Vision Library, [4]), provide numerous algorithms that enable rapid development of visual quality inspection systems and simplify software selection, however, specific details of image processing chain for any particular visual inspection must be defined by an image processing expert on a case-by-case basis, especially if both robust and flexible solution is required.

To eliminate the need for the image processing expert state-of-the art image processing algorithms are bundled together into specialized end-user configurable measurement tools that can be easily set-up. Such tools should be plugins or modules of a larger application where image acquisition, image display, user interface and process control parts are shared/reused. Usually a graphical user interface is used where flowcharts depict image processing pipeline, e.g. NI Vision Builder [5]. Such environments provide many state-of-the art image processing algorithms that can be chained together or are pre-assembled into measurement tools, but there is no universally accepted way of consistently addressing problems related to variations in product position, orientation and size in the acquired image. To further reduce the need for an image processing expert a registration step that would remove variation due to product position, orientation and size should be introduced. Included registration step would enable simpler deployment of more robust image processing solutions in discrete manufacturing[1].

In this paper we propose an *acquire-register-analyze* image processing pattern that is especially suited to discrete manufacturing. Proposed *acquire-register-analyze* pattern aims to increase reproducibility of an image processing flowchart by consistently addressing variations in product position, orientation and size through the registration step that is implemented in a way transparent to the end-user.

---

[1]Discrete manufacturing is production of any distinct items capable of being easily counted.

## II. Image Processing Patterns

For discrete manufacturing once an image is acquired processing is usually done in two steps [6], first is object localization that is followed by object scrutiny and measurement. More detailed structure of an image processing pipeline is given in [1], where typical image processing steps are listed:

1) image acquisition,
2) image preprocessing,
3) feature or object localization,
4) feature extraction,
5) feature interpretation,
6) generation of results, and
7) handling interfaces.

### A. Acquire-Analyze Pattern

Seven typical image processing steps can be decomposed as follows: Image preprocessing step includes image correction techniques such as illumination equalization or distortion correction. Today this is usually done by the acquisition device itself, e.g. GenICam standard [3] requires tools that are sufficient for intensity or color correction. Feature or object localization is a first step of the image processing chain and is usually selected on a case-by-case basis to adjust for variations is object positioning. Feature extraction utilizes typical processing techniques such as edge detection, blob extraction, pattern matching and texture analysis. Results of the feature extraction step are interpreted in the next step to make actual measurements that are used to generate results and to handle interfaces of an assembly line. We call this image processing pattern *acquire-analyze* as feature or object localization must be repeated for (almost) every new feature of interest in the image[2]. A structure of this pattern is shown in Fig. 1.

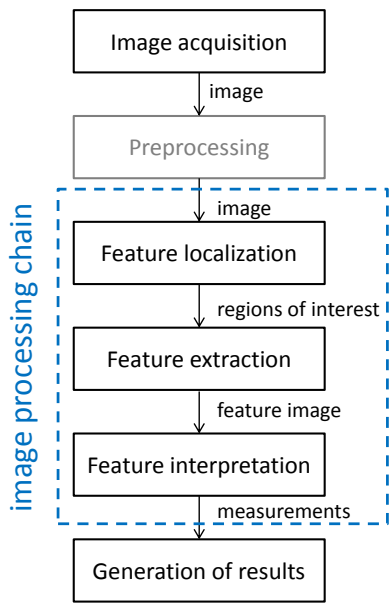

Fig. 1. Acquire-analyze image processing pattern.

---

[2]This holds for discrete manufacturing. For other typed of manufacturing processes registration is not always necessary, e.g. in fabric production feature of interest is texture making registration unnecessary.

Shortcomings of *acquire-analyze* pattern in discrete manufacturing are: a) image processing elements must be chosen and chained together so variation in product placement does not affect the processing, b) processing can be sensitive to camera/lens repositioning or replacement during lifetime of the assembly line, and c) for complex inspections image processing chain requires an image processing expert. Processing can be more robust and the image processing chain easier to define if a registration step is introduced in a way that makes results of feature localization less dependent on variations in position.

When introducing a registration step several requirements must be fulfilled. Firstly, registration must be introduced in a way transparent to the end user, and, secondly, registration must be introduced so unneeded duplication of the image data is avoided.

### B. Acquire-Register-Analyze Pattern

A common usage scenario for visual inspection software in discrete manufacturing assumes the end-user who defines all required measurements and their tolerances in a reference or *source* image of the inspected product. Inspection software then must automatically map the image processing chain and required measurements from the reference or *source* image to the image of the inspected product, the *target* image, by registering two images [7]. End result of the registration step is: a) removal of variation due to changes in position of the product, b) no unnecessary image data is created as no image transforms are performed, instead image processing algorithms are mapped to the *target* image, c) as only image processing algorithms are mapped overall mapping can be significantly faster then if the *target* image data is mapped to the *source* image, and d) no image interpolation is needed which can lead to overall better performance as any interpolation artifacts are eliminated by the system design.

In discrete manufacturing all products are solid objects so feature based global linear rigid transform is usually sufficient to map the defined image processing chain from the *source* to the *target* image; that is, compensating translation, rotation and scaling by using simple homography is sufficient. Required mapping transform is defined by a $4 \times 4$ matrix $\mathbf{T}$ [8]. This transform matrix must be transparently propagated through the whole processing chain. If a graphical user interface for defining the processing chain is used this means all processing tools must accept a transform matrix $\mathbf{T}$ as an input parameter that can be optionally hidden from the end user to achieve transparency.

We call this image processing pattern *acquire-register-analyze* as feature or object localization is done once at the beginning of the pipeline to find the transform $\mathbf{T}$. A structure of this pattern is shown in Fig. 2.

## III. Software Design and Implementation

To test the viability and usefulness of the proposed *acquire-register-analyze* pattern a visual inspection software was constructed. Software was written in C++ and C# programming languages for the .Net platform and is using Smartek GigEVision SDK [9] for image acquisition, and OpenCV [4] and Emgu CV [10] for implementing image processing tools.

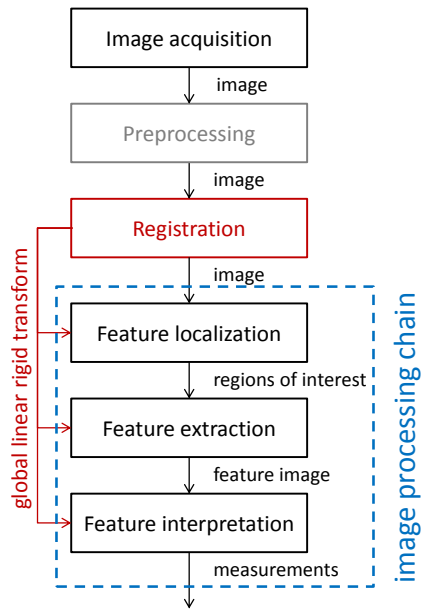

Fig. 2. Acquire-register-analyze image processing pattern.

Graphical user interface (Fig. 3) and logic to define an image processing chain was written from scratch using C#.

*A. User Interface*

Interface of the inspection software must be designed to make various runtime task simple. Typical runtime tasks in visual inspection are [1]: a) monitoring, b) changing the inspection pipeline, c) system calibration, d) adjustment and tweaking of parameters, e) troubleshooting, f) re-teaching, and g) optimizing. Those various tasks require several different interfaces so the test application was designed as a tabbed interface with four tabs, first for immediate inspection monitoring and tweaking, second for overall process monitoring, third for system calibration and fourth for definition of the image processing chain (Fig. 3).

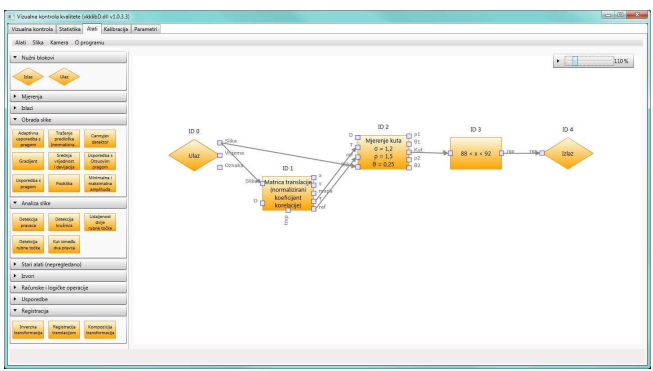

Fig. 3. User interface for defining the image processing chain by flowchart.

User interface for defining an image processing tool-chain must be composed so the end-user is able to intuitively draw any required measurement in the reference image. Interface should also hide most image processing steps that are required to compute the measurement and that are not the expertise of the end-user, e.g. edge extraction, ridge extraction, object segmentation, thresholding, blob extraction etc. This is achieved by representing all measurements by rectangular blocks. Inputs are always placed on the left side of the block while outputs are always placed on the right side of the block. Name is always shown above the block and any optional control parameters are shown on the bottom side of the block. Several typical blocks are shown in Fig. 4.

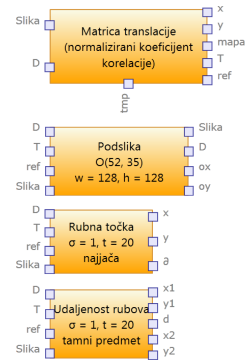

Fig. 4. Example for four image processing tools. Inputs are always on the left side, outputs are always on the right side and optional parameters are on the bottom of the block.

*B. Registration*

Two blocks that are present in every inspection flowchart are input and output blocks. For the *acquire-register-analyze* pattern first block in the flowchart following the input block is a registration block that internally stores the reference or *source* image. For every *target* image the transform is automatically propagated to all following blocks. Note that actual implementation of the registration is not fixed, any technique that enables recovery of simple homography is acceptable. Using OpenCV [4] a simple normalized cross-correlation or more complex registration based on key-point detectors such as FAST, BRISK or ORB can be used.

Figs. 5 and 6 show the *source* and the *target* images with displayed regions of interest where lines are extracted for the angle measurement tool. User can transparently switch between the *source* and the *target* as the software automatically adjusts the processing tool-chain.

*C. Annotations and ROIs*

One additional problem when introducing registration transparent to the end-user are user defined regions-of-interest (ROIs) and annotations that show measurement and inspection results on a non-destructive overlay.

Every user-defined ROI is usually a rectangular part of the reference or *source* image that has new local coordinate system that requires additional transform matrix so image processing chain can be mapped first from the *source* to the *target*, and then from the *target* to the ROI. Composition of transforms achieves the desired mapping. However, as all image processing tools are mapped to the input image or to defined ROIs and as there can be many different mapping transforms the inspection result cannot be displayed as a non-destructive overlay unless a transform from local coordinate system to the *target* image is known[3]. This transform to achieve correct overlay display can also be described by a $4 \times 4$ matrix $\mathbf{D}$ that must be computed locally and propagated every time a new ROI is introduced in the image processing chain.

So for the *acquire-register-analyze* pattern elements of the image processing chain can be represented as blocks that must accept $\mathbf{T}$ and $\mathbf{D}$ as inputs/outputs that are automatically connected by the software (Fig. 4), and that connections can be be auto-hidden depending on the experience of the end-user.

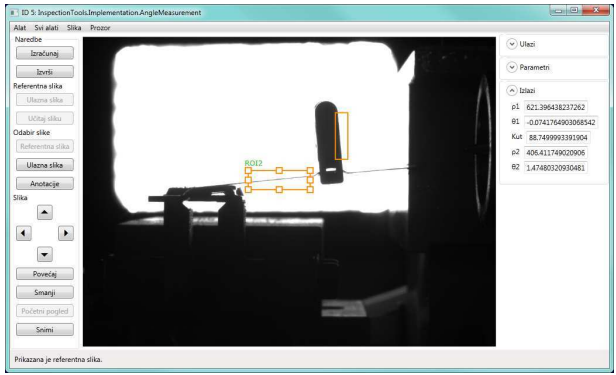

Fig. 5.  Tool's ROIs for line extraction and angle measurement shown in the *source* image.

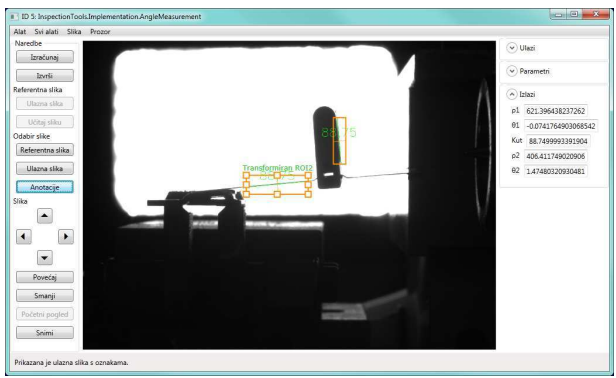

Fig. 6.  Tool's ROIs for line extraction and angle measurement shown in the *target* image.

## IV. RESULTS

Developed prototype software using the proposed *acquire-register-analyze* pattern is in production use at the Elektro-Kontakt d.d. Zagreb plant. The plant manufactures electrical switches and energy regulators for electrical stoves (discrete manufacture). Observed variations in acquired images that must be compensated using registration are caused by product movement due to positioning tolerances on a conveyor and by changes in a camera position[4].

Regardless of the cause the variation in a product placement with respect to the original position in the *source* image can be computed. Thus every used image processing chain starts with a registration block that registers the prerecorded *source* image to the *target* image and automatically propagates the registration results through the processing chain.

---

[3]Note that this is NOT a transform back to the *source* image.

[4]After regular assembly line maintenance camera position is never perfectly reproducible.

Developed prototype software was accepted by the engineers and staff in the plant and has successfully completed the testing stage and is now used in regular production.

### A. Contact Alignment Inspection

There are two automated rotary tables with 8 nests for welding where two metal parts are welded to form the spring contact. There are variations in product placement between nests and variations in the input image due to differences in camera placement above the two rotary tables (see Fig. 8).

Image processing chain is composed of the registration step that automatically adjusts position of subsequent measurement tools that measure user defined distances and angles and compare them to the product tolerances. This proposed design enables one product reference image to be used where required product dimensions are specified and that is effectively shared between production lines thus reducing the engineering overhead as only one processing pipeline must be maintained.

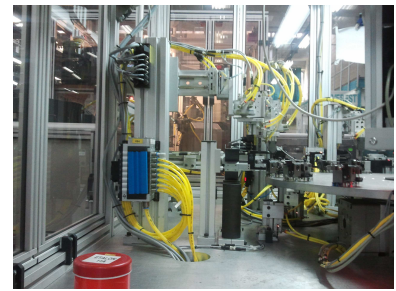

Fig. 7.  Rotary table for contact welding.

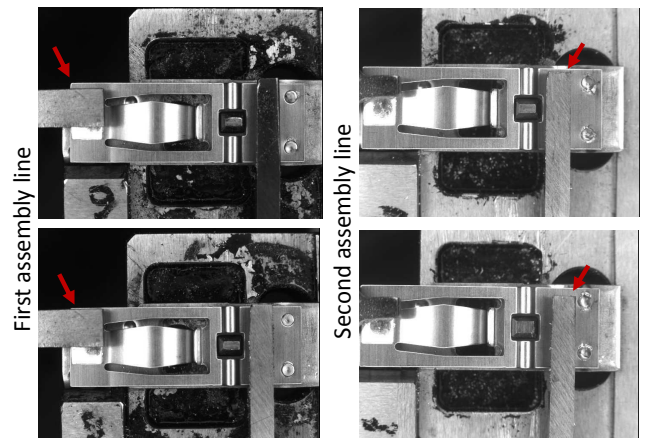

Fig. 8.  Variations in product placement.

### B. Energy Regulator Inspection

There are four assembly lines where an energy regulator for electrical stoves is assembled. At least seventeen different product subtypes (Fig. 3) are manufactured. There is a variation in product placement due to positioning tolerances on the conveyor belt and due to variations in cameras' positions across all four assembly lines. Defined inspection tool-chains must be shareable across the lines to make the system maintenance simple.

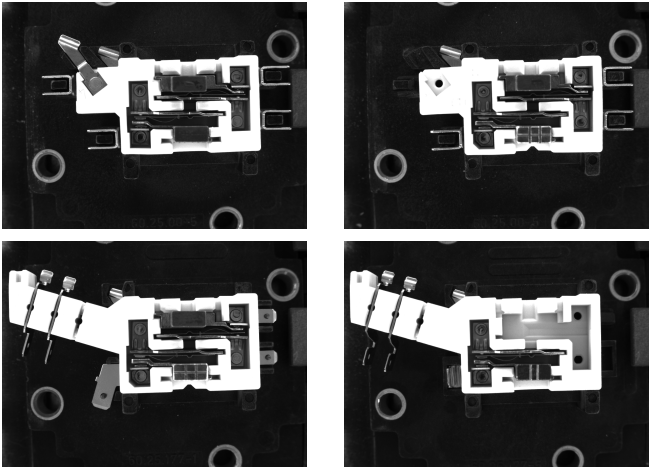

Fig. 9. Four different product variants (body and contact shape) of an energy regulator recorded on the same assembly line.

Large number of product variants requires the user interface that enables a flexible adaptation and tweaking of the processing pipeline. Here the graphical programming environment (Fig. 3) was invaluable as it enables the engineer to easily redefine the measurements directly on the factory floor. Furthermore, due to the registration step only one reference image per product subtype is required so only seventeen different image processing chains must be tested and maintained.

## V. Conclusion

In this paper we have presented an *acquire-register-analyze* image processing pattern that aims to increase the robustness of an image processing flowchart by consistently addressing variations in product position, orientation and size. Benefits of the proposed pattern are: a) no unnecessary image data is created from the input image by the application during image processing, b) overall speed and throughput of the image processing pipeline is increased, and c) interpolation artifacts are avoided.

We have demonstrated the feasibility of the proposed pattern through case-studies and real-world use at the Elektro-Kontakt d.d. Zagreb plant.

## Acknowledgment

The authors would like to thank Elektro-Kontakt d.d. Zagreb and mr. Ivan Tabaković for provided invaluable support.